\documentclass[11pt]{article}

\usepackage[margin=1in]{geometry}
\usepackage{graphicx}
\usepackage{amsmath}
\usepackage{amssymb}
\usepackage{authblk}
\usepackage[hidelinks]{hyperref}
\usepackage{natbib}
\usepackage{booktabs}
\usepackage{caption}
\title{\bfseries A scaling law of contextual persistence in human language}

\author[1]{Elan Barenholtz}
\affil[1]{Machine Perception and Cognitive Robotics Laboratory, Department of Psychology and
Center for Complex Systems, Florida Atlantic University}

\date{}

\begin{document}

\maketitle

\begin{abstract}
\noindent
Human language exhibits lawful structure at the level of words (frequency,
vocabulary growth) and word pairs (co-occurrence across distance). Here we show the arrangement of words in
sequence --- a central determinant of meaning --- obeys a comparable law. Using
large language models as probes, we measured the reduction in target perplexity
conferred by context at distance $d$ beyond that of the same words scrambled;
this difference, the contextual persistence function $P(d)$, isolates the influence of
arrangement. Across ten written and spoken corpora spanning six
language families, $P(d)$ decayed approximately as $1/d$
($P(d)\propto d^{-\alpha}$, mean $\alpha=1.04$; median $r^2=0.96$). The effect vanished
in scrambled and synthetic controls, replicated across independent probes, and its scale-free
form was absent from genomic and protein sequences under domain-native models. An exponent near $1$ distributes
contextual influence approximately uniformly across logarithmic timescales. The results
establish a scaling law of contextual persistence in human language --- a quantitative
regularity of communicative behaviour.
\end{abstract}

\section{Introduction}

Human language has a rich statistical structure, aspects of which have been captured by quantitative
laws \citep{altmann2016statistical, tanakaishii2021statistical}. Word frequencies follow Zipf's law \citep{zipf1949human}, and
vocabulary growth follows Heaps' law \citep{heaps1978information}; both are scale-free and recur
with striking consistency across languages and corpora. What makes such a regularity a law is the
reproducibility of its exponent, not the power-law form alone: across a wide range of languages and
corpora, Zipf's exponent sits near unity and Heaps' recurs within a characteristic sublinear band.
Both of these laws describe the units in aggregate, and so occupy
the most basic level of a hierarchy of statistical structure. A second level concerns
the units not singly but in pairs: how their occurrences co-vary across distance. Here too a high
degree of statistical structure has been uncovered: the mutual information
between symbols falls off as an approximate power law of their separation, decaying far more slowly
than expected under conventional finite-order Markov models \citep{ebeling1994entropy, lin2017critical, montemurro2002longrange}.

Neither of these measures, however, isolates the order in which units appear. Unit frequencies
are order-blind by construction, and long-range mutual information can reflect repeated lexical
and topical content even when the local ordering of that content is
disrupted. Yet the same words at the same separations can compose entirely different meanings
depending on their arrangement: grammatical relations and compositional meaning live in
arrangement itself, not in the inventory of words or their spacing. Representational accounts of
contextual dependence, such as discourse and situation models \citep{vandijk1983strategies} and
resonance-based accounts of memory access during reading \citep{myers1998accessing}, describe it
qualitatively; no lawful quantitative structure has been identified. Like co-occurrence,
order-specific dependence can be posed as a question of distance: how does the probability of a
given word depend on the arrangement of the other words in the sequence, and how does that
dependence change with distance?

Reaching this third level requires estimating the probability of specific words given ordered
combinations of other words. This is the arrangement-level analogue of the pairwise calculation
underlying long-range mutual information: where mutual information asks how the occurrence of one
word co-varies with that of another at a given separation, this quantity asks how the probability
of a word depends on the ordered configuration of others --- whether immediately adjacent, as in
what follows a given construction, or far back in the sequence. Classical corpus statistics
approximate it only crudely. Autoregressive transformer models
\citep{vaswani2017attention, radford2019language} provide a tool for capturing this structure
directly: trained only to predict the next token, they assign token-level conditional
probabilities over context lengths far exceeding those tractable with conventional corpus
estimators, and are sensitive to precisely the order-dependent structure that raw co-occurrence
statistics cannot isolate. Used as measuring instruments, or probes, rather than as objects of study, they provide a
new vantage point on this level of linguistic structure.

Here we measure this dependence directly. Like mutual information, the quantity is a statistical
dependence, but it is taken between an individual word and the ordered configuration of the words
that precede it rather than between pairs of words. A probe's next-token probability supplies the
measure of influence: how much the probability assigned to a fixed upcoming span changes when prior
context is revealed. Distance along the sequence provides the axis of measurement: revealing the true
context step by step, farther and farther back, makes that influence a function of distance $d$.

The method of measuring next-token prediction as a function of context length has a distinguished
pedigree. Shannon's guessing experiments
\citep{shannon1951prediction} asked how much uncertainty about what comes next remains as a reader
is given longer and longer stretches of the preceding text, and Hilberg's conjecture
\citep{hilberg1990grenzwert, debowski2015relaxed}, a reinterpretation of Shannon's data, holds
that the answer follows a power law: prediction keeps improving with context length, with no
floor reached within the measured range. Our intact-context condition extends this measurement,
with a probe reaching far longer contexts. Critically, unlike these earlier studies, our measure
specifically addresses the role of order within the sequence. In that tradition context was always
presented intact and only its length was varied, so the resulting curve is a total, mixing what is
gained from the words revealed with what is gained from their arrangement, with no way to tell the
two apart. At
each step we therefore ask how much the revealed context reduces the probe's surprise, and subtract the
reduction obtained when the same words are supplied in scrambled order --- the entire revealed span
permuted, so that the baseline retains vocabulary and topical content but no arrangement. What
remains is the additional benefit of intact sequential arrangement: the order-specific contextual
influence. Its per-token gain as context is extended past distance $d$ defines the
\emph{contextual persistence function} (CPF), $P(d)$; its functional form we call the
\emph{contextual persistence law}. Note: we use \emph{persistence} in a statistical sense --- the
continued predictive influence of ordered prior context as a function of distance --- not the
persistence of any internal representation or cognitive process: the function is the
operational measure, and the law is the regularity it reveals. We apply it across ten
corpora spanning six language families, multiple genres, and written and spoken modalities.
Throughout, we treat the CPF as a law governing natural-language sequences, a property of
the linguistic product --- a statistical trace of human communicative behaviour; whether and how it maps onto the real-time mechanisms of human
production and comprehension is a separate, empirically open question we return to in the
Discussion.

We find that $P(d)$ decays approximately as $1/d$ ($P(d)\propto d^{-\alpha}$), with mean decay exponent
$\alpha=1.04$ (SD $=0.15$) and no characteristic cutoff across nearly two orders of magnitude in distance. This decay
is consistent across language families, genres, and modalities, vanishes in
random-token controls, scales with the integrity of sequential structure, and is
broadly distributed across prior sentences rather than concentrated in a small number
of anchors. The CPF describes a continuous, scale-free temporal
structure in human language, measurable directly from the linguistic sequence and
replicable across independently trained probe models
(Llama-3.1-8B, Mistral-7B).

The shape of $P(d)$ places language alongside other human behaviours and neural
processes that exhibit scale-free temporal structure \citep{gilden2001cognitive, barabasi2005origin, linkenkaerhansen2001longrange, beggs2003neuronal, kello2010scaling}, and constrains the class of
mechanisms capable of producing such structure. In particular, the absence of any
characteristic scale across the measured range, combined with the graded
contributions of prior content distributed across the sequence, is difficult to
reconcile with architectural accounts of memory that posit a sharp division between
maintained and retrieved material with distinguishable temporal dynamics. The pattern
is consistent instead with mechanisms in which prior content continues to shape
ongoing generation through the evolving probability structure of the sequence itself.

\section{Results}

\subsection{The contextual persistence function}

We define the contextual persistence function (CPF), $P(d)$, as the \emph{order-specific}
predictive influence of prior context at distance $d$ from the current point of generation:
the per-token reduction in target perplexity produced by extending the revealed context past
distance $d$, minus the corresponding marginal computed with the entire revealed span in scrambled
order (Methods). Subtracting the
shuffled-token baseline controls the predictive benefit available from the same token composition in
scrambled order, so $P(d)$ isolates the additional contribution of intact sequential arrangement,
resolved as a function of distance.

\begin{figure}[t]
\centering
\includegraphics[width=\textwidth]{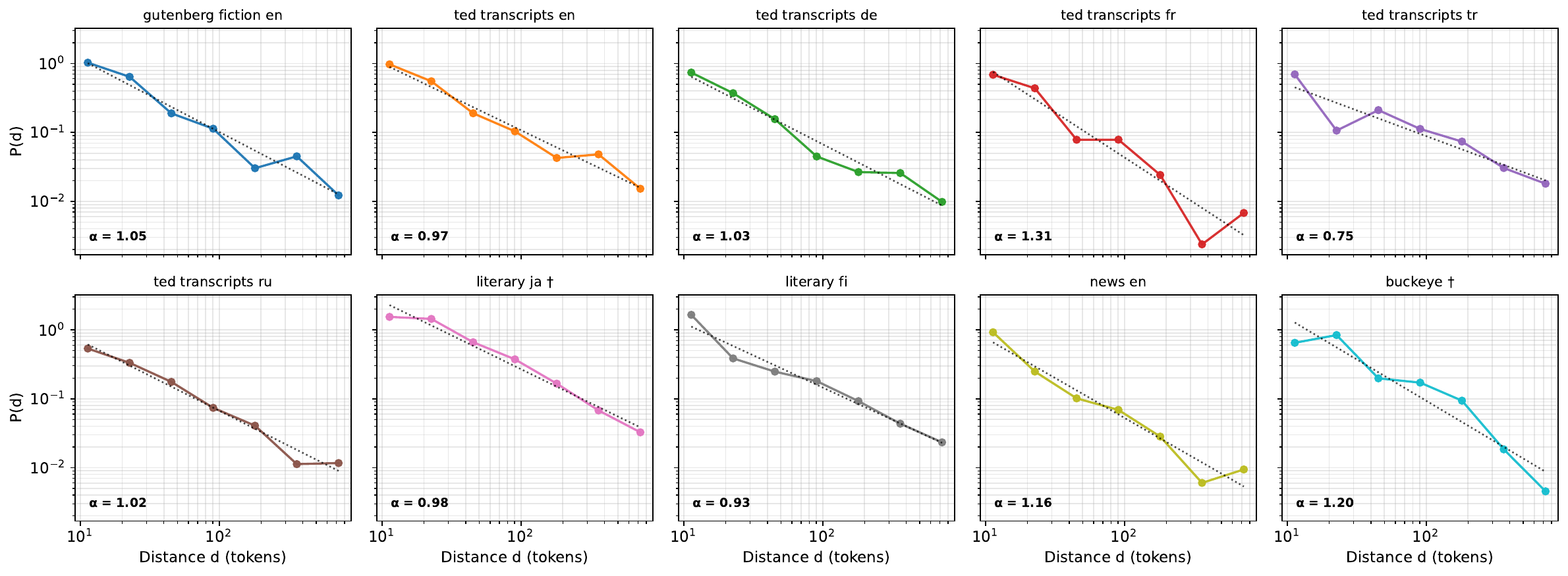}
\caption{\textbf{Cross-corpus contextual persistence functions.} Per-token order-specific contextual
influence $P(d)$ as a function of distance $d$, log--log axes. Each panel is one of ten corpora spanning
six language families (Germanic, Romance, Turkic, Slavic, Japonic, Uralic) and four production modes. All
cells exhibit heavy-tailed decay across nearly two orders of magnitude in distance, with mean decay
exponent $\alpha=1.04$ (SD $=0.15$; $P(d)\propto d^{-\alpha}$, so the log--log slope is $-\alpha$) ---
close to a $1/d$ law. Points are shown for $d \geq 10$ tokens; the dotted line is the power-law fit. Intervals below $d = 10$ are excluded because at such short context the
shuffled-token baseline is unstable: with only one or two tokens there is no meaningful permutation, so
the order-specific difference is ill-defined there and does not reflect the persistence decay (Methods).
The two cells marked $\dagger$ (Japanese literary prose; Buckeye spontaneous English speech) show mild
downward curvature and are marginally better fit by a log-normal and a stretched-exponential form
respectively, but remain heavy-tailed with the power law a close second.}
\label{fig:persistence}
\end{figure}

Figure~\ref{fig:persistence} shows $P(d)$ for each of the ten corpora, plotted against distance
on log--log axes. Across all corpora, $P(d)$ is positive throughout the measured range: at
distances approaching 1000 tokens (the far end of our measurement range), prior context continues
to reduce target perplexity by a measurable margin relative to shuffled controls. Long-range
contextual influence is not confined to a recency window but extends as a graded function across
the full sequences we examined.

The analyses that follow characterize this function: its form, a power law with exponent near
$1$ that recurs across languages, genres, and modalities; its source, sequential structure in
language itself rather than properties of the probe, established by synthetic, cross-probe, and
cross-domain controls; and its composition, the separable contributions of sentence-internal
order, discourse sequence, and sentence content, distributed broadly across prior context rather
than concentrated in privileged sentences.

\subsection{Heavy-tailed decay across languages, genres, and modalities}

The functional form of $P(d)$ is well approximated by a heavy-tailed decay. Across
languages, genres, and modalities, $P(d)$ decays smoothly across nearly two orders of
magnitude with a mean decay exponent $\alpha=1.04$ (SD $=0.15$; $P(d)\propto d^{-\alpha}$)
--- close to a $1/d$ power law --- and a median goodness-of-fit of $r^2 = 0.96$. A
reduced-sample replication averaging $20$ shuffled permutations per target produced the same
cross-corpus mean exponent ($\alpha=1.04$) and improved fit stability (Supplementary Note~3),
indicating that the headline result does not depend on the selected permutation. Fits are computed on $d \geq 10$
tokens, the range over which both ordered- and shuffled-context bases are stable;
very-short-range intervals are reported in Methods, and the fitted exponent is insensitive to the
exact cutoff within the stable range (Supplementary Note~4). In a comparison of five candidate
functional forms by small-sample-corrected information criteria (Supplementary Note~1), the power law is preferred in eight of the ten corpora and outperforms a
pure exponential in all ten; a fixed-scale exponential is never preferred. The two
remaining corpora, Japanese literary prose and Buckeye spoken English, show mild downward
curvature and are marginally better described by a log-normal and a stretched-exponential
form respectively; in both the decay remains heavy-tailed and extends well beyond the range
expected from a fixed temporal buffer acting alone.

The cross-corpus consistency of the decay regime is notable. The sample spans
Germanic, Romance, Turkic, Japonic, Uralic, and Slavic language families, as well as
fiction and literary prose, news and expository prose, prepared speech transcripts,
and spontaneous speech. A narrow heavy-tailed regime across this range indicates that
the CPF reflects a general property of natural language rather than a
property of any single language, register, or modality.

The qualitative pattern that defines $P(d)$ --- a positive contribution of ordered
context at long range together with a near-zero contribution of shuffled context ---
extends beyond the six families covered by the long-range run. In a complementary
cross-language analysis evaluating shorter context lengths on both probe models, the
long-range sign pattern is preserved across three additional language families:
Afro-Asiatic, Koreanic, and Sino-Tibetan (Supplementary Information). The headline
exponent of $P(d)$ is therefore obtained on a six-family set within a broader nine-family
cross-language replication of the qualitative pattern that defines the persistence
function.

\subsection{An exponent near $1$: uniform influence across logarithmic timescales}

Two properties of the decay should be separated. The first is that it is scale-free: a power
law has no distance at which its behaviour changes character, and our fits show none across
the measured range. The second, and more distinctive, is the value of the exponent. Across the
ten corpora the fitted decay exponent is close to $1$ and clustered around it (mean $\alpha=1.04$,
SD $=0.15$, with $P(d)\propto d^{-\alpha}$).
The observed between-corpus spread modestly exceeds document-sampling error (typical bootstrap SE
${\approx}\,0.10$; Supplementary Note~2), although permutation sensitivity analyses indicate that
part of this spread reflects noise in the shuffled baseline. The exponent is therefore best described
as clustered near $1$, with limited evidence for residual corpus-level heterogeneity. The reproducibility of the near-unity value across otherwise very different samples is
itself part of the result, and it carries a specific quantitative meaning.

\begin{figure}[t]
\centering
\includegraphics[width=\textwidth]{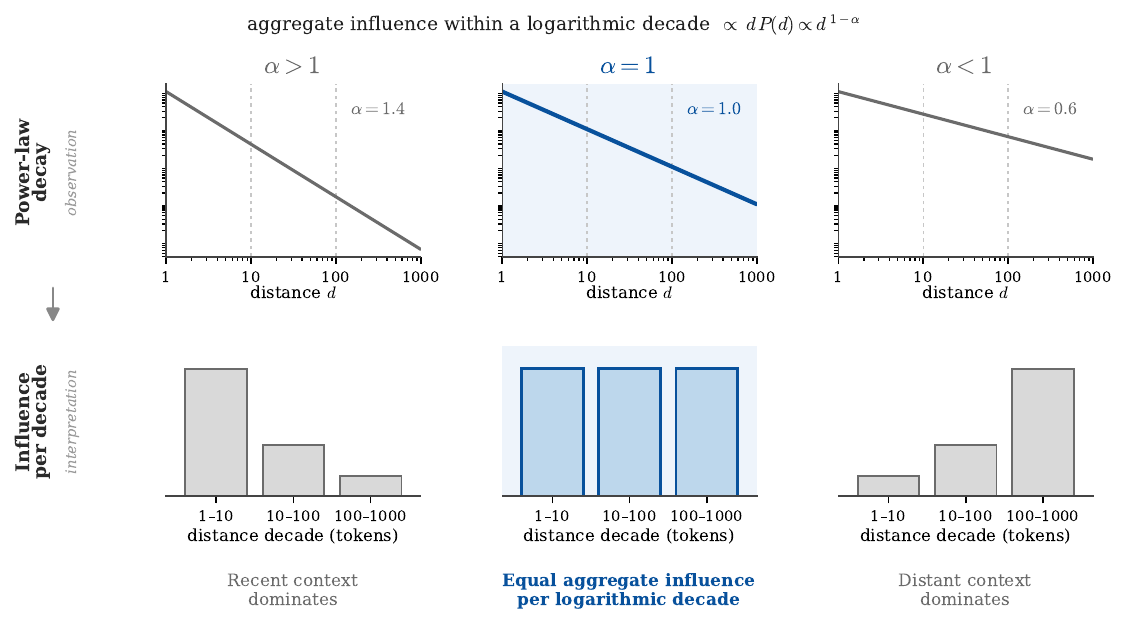}
\caption{\textbf{Persistence functions $P(d)\propto d^{-\alpha}$ at three exponents and the resulting per-decade influence (schematic).}
\textbf{Top row (observation):} the persistence law $P(d)\propto d^{-\alpha}$ on log--log axes, where the
decay appears as a straight line of slope $-\alpha$. The three power laws look nearly indistinguishable.
\textbf{Bottom row (interpretation):} their sharply different consequences. The aggregate influence within a
logarithmic decade of distance is proportional to $d\,P(d)\propto d^{\,1-\alpha}$, so for $\alpha>1$ (left)
aggregate influence is concentrated in recent context; for $\alpha<1$ (right) it is dominated by distant
context; and at $\alpha\approx1$ (centre) each logarithmic decade of the past --- 1--10, 10--100,
100--1{,}000 tokens back --- contributes approximately equally, distributing influence uniformly across
logarithmic timescales. The measured mean exponent ($\alpha=1.04$) places human language at this
boundary. The lower bands are a conceptual illustration, not data.}
\label{fig:alpha}
\end{figure}

For a CPF of the form $P(d) \propto d^{-\alpha}$, the aggregate contextual influence contained
within a logarithmic interval of distance --- the band spanning one multiplicative step, or one
decade, of the past --- is proportional to $d\,P(d) \propto d^{\,1-\alpha}$. When $\alpha \approx 1$
this quantity is approximately constant. Put concretely, if one divides the past into logarithmic
bands (1--10, 10--100, and 100--1{,}000 tokens back), each band contributes approximately the
same aggregate contextual influence, even though the influence of any single word continues to
decline with distance. The reason the decades flatten is a near-exact cancellation of two opposing
effects. Each successive logarithmic band reaches proportionally further back and so contains
proportionally more words, and at $\alpha \approx 1$ that growth in number almost exactly offsets
the $1/d$ decline in the average influence per token, so the many weakly contributing words far
back combine to match the few strongly contributing words nearby. Contextual influence is therefore
distributed approximately uniformly across
logarithmic timescales: no logarithmic timescale dominates the aggregate, and every decade of the
past contributes comparably to ongoing prediction (Figure~\ref{fig:alpha}).

This should not be read as memory being equally strong at all distances; individual words do
become progressively less influential as they recede. What is approximately equal is the
\emph{aggregate} influence summed within each logarithmic decade: pointwise influence falls as
$1/d$, while cumulative influence per multiplicative interval of distance remains flat.

The value $\alpha \approx 1$ is also a boundary. For $\alpha > 1$, aggregate influence is
concentrated near the present and the remote past contributes negligibly in sum; for $\alpha < 1$,
the sum is increasingly dominated by distant context. $\alpha \approx 1$ is the crossover between
these regimes: the exponent at which no timescale dominates the total. That human language sits
at this boundary, rather than at an arbitrary point along the continuum, is what most distinguishes
the result from a generic heavy tail.

The near-unity regime is robust to the choice of response scale: expressing influence in additive
log-probability (nats) units rather than perplexity leaves every corpus a near-$1/d$ power law,
though the mean exponent shifts modestly from $\alpha=1.04$ to $\alpha=1.18$ (Supplementary Note~2), indicating approximate rather than exact equality of influence across logarithmic
decades.

The reproducibility of the near-unity exponent across corpora is what supports treating this as a
candidate scaling law rather than an isolated curve fit. As with Zipf's law, the recurrence of the
value --- an exponent near $1$ across languages, modalities, and independently trained probe models
--- is the phenomenon, not merely the power-law form.

\subsection{The CPF reflects sequential structure, not properties of the probe}

A central concern in any LLM-based measurement is whether the observed pattern
reflects properties of natural language or properties of the probe model. We address
this directly with synthetic-sequence controls. We constructed sequences with
token-frequency statistics matched to natural text but with sequential structure
removed. Raw perplexity over these synthetic sequences exhibits apparent power-law
structure with a slope comparable in magnitude to that observed on natural text
(Llama-3.1: $\approx -1.27$ on uniform random vocabulary; Mistral-7B: $\approx -1.34$),
replicating the general tendency for language-model perplexity to fall as context
length increases even when genuine long-range dependence is absent.

The order-specific CPF, $P(d)$, however, is near zero across all
distances on the synthetic sequences. The shuffled-token baseline that defines $P(d)$
absorbs the perplexity reduction attributable to token statistics alone, isolating the
contribution of sequential organization. The long-range, heavy-tailed $P(d)$ observed
for natural language therefore depends on properties of the linguistic sequence
itself, not on the probe. We replicated the persistence signature with two
independently trained probe models, Llama-3.1-8B and Mistral-7B, obtaining the same
qualitative pattern and comparable decay ranges on shared corpora, along with near-zero
$P(d)$ on synthetic controls. The CPF is therefore not an idiosyncratic
property of the model used to estimate it. Both probes do share the transformer architecture;
replication with architecturally distinct autoregressive models (for example, state-space models)
remains an additional check for future work.

\begin{figure}[t]
\centering
\includegraphics[width=\textwidth]{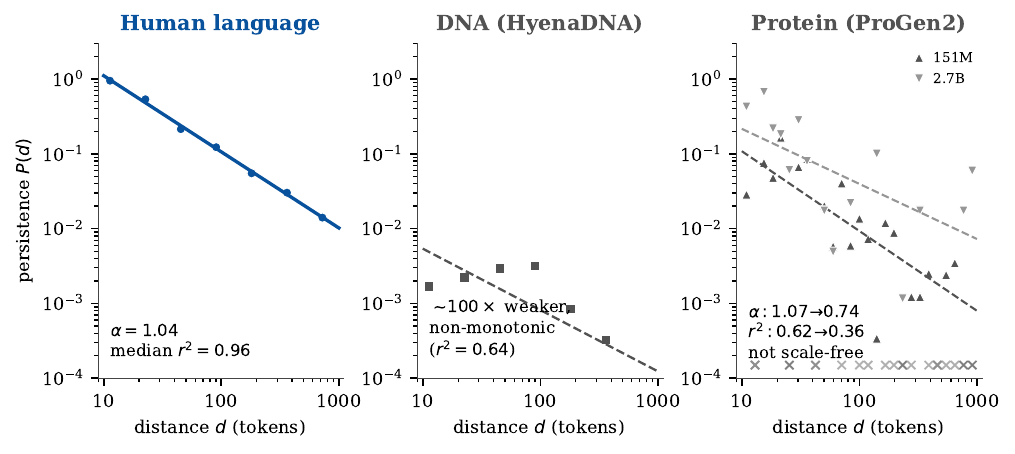}
\caption{\textbf{Order-specific persistence $P(d)$ for human language, DNA, and protein under the
identical protocol.} Order-specific persistence $P(d)$ under the identical CPF protocol, on log--log axes,
for human language (aggregate of the ten corpora), DNA (HyenaDNA), and protein (ProGen2 at two
scales, 151M and 2.7B). \textbf{Left:} language decays as a clean power law (mean $\alpha=1.04$,
median $r^2=0.96$; points lie on the fit). \textbf{Centre:} DNA is roughly two orders of magnitude
weaker and non-monotonic, with no clean scale-free decay. \textbf{Right:} protein carries genuine long-range
influence but is not scale-free --- the fit is poor and the point estimate varies across model
scale while power-law fit quality declines ($\alpha=1.07\!\to\!0.74$, $r^2=0.62\!\to\!0.36$ from 151M to 2.7B); $\times$ marks distances
where $P(d)\leq 0$. The discriminating combination is scale-freeness and exponent stability, not the
mere presence of long-range structure.}
\label{fig:langbio}
\end{figure}

A stronger version of this argument replaces synthetic sequences with real non-language
sequences that carry their own long-range structure, each probed with its own domain-native
autoregressive model. If the law were a by-product of autoregressive probing rather than a
property of language, the identical CPF protocol should reproduce it elsewhere. It does not
(Figure~\ref{fig:langbio}).
On genomic sequence (human chromosome~21, HyenaDNA), $P(d)$ is near zero, non-monotonic, and
far weaker than language, with no scale-free decay. On protein sequence (Swiss-Prot, ProGen2),
there is genuine long-range influence (expected from coevolutionary structure) but not
language's form: the power-law fit is poor, and the point-estimate exponent varies across model
scale while the fit itself deteriorates ($r^2 = 0.36$--$0.62$ versus $0.96$ for language;
Supplementary Information), so the curve is long-range but not scale-free. Long-range dependence
is not unique to language. The distinguishing result is the combination of scale-free decay, a
near-unity exponent, and cross-probe consistency: among the domains and probes tested, only human
language shows all three, with the measurement protocol and the presence of long-range structure
held in common. The regularity is a property of the linguistic sequence, not of the instrument.

\subsection{Sequence, reversal, and content make distinguishable contributions to persistence}

\begin{figure}[t]
\centering
\includegraphics[width=0.85\textwidth]{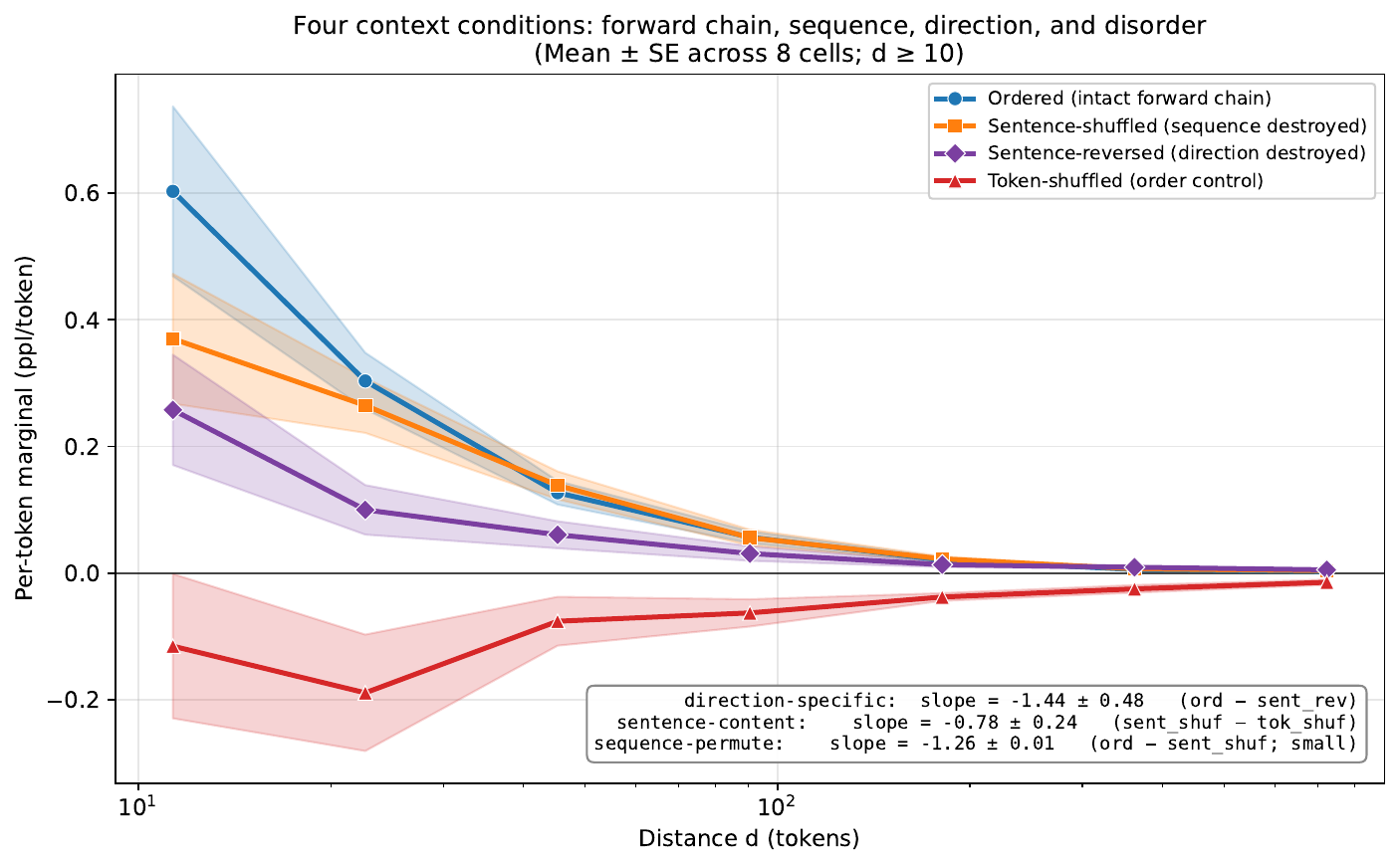}
\caption{\textbf{Per-token order-specific marginals under four context conditions.}
Per-token order-specific marginals for four context conditions on log--log axes, aggregated across
eight cells: ordered (intact forward chain), sentence-shuffled (discourse sequence destroyed, sentence
content preserved), sentence-reversed (sentence order reversed, so the set of adjacencies is preserved
while their order and target-relative positions are reversed), and token-shuffled (ordered structure
removed). Curve ribbons are mean $\pm$ SE across the eight cells. Three contrasts decompose $P(d)$:
sentence-reversal (intact minus sentence-reversed; slope $-1.44 \pm 0.48$), sequence-permutation (intact
minus sentence-shuffled; slope $-1.26 \pm 0.01$), and sentence-content (sentence-shuffled minus
token-shuffled; slope $-0.78 \pm 0.24$); slope uncertainties are the SD of the per-cell slopes. The
sentence-reversal and sequence-permutation contrasts are concentrated below ${\sim}100$ tokens, whereas
the sentence-content contrast is shallow and carries the long-range tail. The reported quantities are
log--log slopes, not values of $\alpha$; since $P(d)\propto d^{-\alpha}$ with $\alpha>0$, a slope of
$-1.44$ corresponds to a decay exponent $\alpha=1.44$.}
\label{fig:sentshuf}
\end{figure}

To separate the effects of disrupting, reversing, and dissolving discourse structure, we evaluated
four context conditions on the same target windows (Figure~\ref{fig:sentshuf}): intact context;
sentence-shuffled context, which preserves sentence content and within-sentence order
while disrupting the discourse sequence; sentence-reversed context, which preserves the
set of sentence adjacencies while reversing their order and target-relative positions; and
token-shuffled context, which removes ordered structure altogether. Three contrasts
follow. The sentence-reversal contrast (intact minus sentence-reversed) measures the loss
associated with reversing the temporal organization of the prior discourse, while preserving
sentence-internal order and the set of sentences; the sequence-permutation contrast (intact minus
sentence-shuffled) isolates sentence-sequence organization; and the sentence-content
contrast (sentence-shuffled minus token-shuffled) estimates the contribution retained by
sentence content once discourse ordering is removed.

Over their positive range, each contrast is approximately linear on log--log axes, with distinct
fitted slopes. The sentence-reversal and sequence-permutation contrasts are steep and
concentrated at short range (slopes $-1.44 \pm 0.48$ and $-1.26 \pm 0.01$
respectively, approaching zero by $d > 100$ tokens), whereas the sentence-content contrast
is shallow and carries the long-range tail (slope $-0.78 \pm 0.24$). Sensitivity to sentence
reversal and discourse-sequence permutation is therefore concentrated at short range, whereas
sentence-level content carries the long-range tail.

\paragraph{A localized order manipulation converges on the same decomposition.}
As an independent check, we applied an increment-shuffle manipulation that
localizes the disruption to a single distance band: for each adjacent pair of
context lengths $(c_{i-1},c_i)$, all nearer context (distances $1\ldots c_{i-1}$)
is held intact and in place, and only the tokens at distances
$c_{i-1}+1\ldots c_i$ are permuted within that band ($K=20$ permutations per
target and band). The per-token quantity
$Q(d)=(\overline{\mathrm{ppl}}_{\mathrm{shuf}}-\mathrm{ppl}_{\mathrm{intact}})/(c_i-c_{i-1})$,
at the geometric midpoint $d=\sqrt{c_{i-1}c_i}$, isolates the internal order of a
single distant span while the surrounding context is preserved; where $P(d)$
reflects the order-specific value of the accumulating context as a whole, $Q(d)$
probes span-internal order alone. Across the eight cells with a range-matched
$P(d)$ estimate (Figure~\ref{fig:qd}), $Q(d)$ is positive throughout but decays more steeply than
$P(d)$, with a log--log slope near $-1.3$ and downward curvature (bounded and
curved forms outperform a pure power law by AIC in most cells), whereas the
range-matched $P(d)$ retains a slope near $-1.0$; the band-local exponent exceeds
the integrated exponent in every cell tested. This coincides with the
sequence-permutation contrast ($-1.26\pm0.01$) recovered from sentence shuffling,
obtained here by an independent manipulation at token resolution with nearer
context held fixed. Two structure-disrupting methods therefore converge: the
internal ordering of localized spans makes a steep, short-range, bounded
contribution, while the scale-free long-range tail is carried by integrated
content rather than by the internal order of any distant region. We note what this contrast does
and does not license. In natural generation no span is conditioned on in isolation; each word is
produced against the entire ordered sequence so far. The extension-marginal $P(d)$ therefore
tracks a quantity with a direct referent in natural generation --- the influence that accrues as
ordered context lengthens --- whereas the increment shuffle is a decomposition tool, informative
precisely because it shows where the scale-free law does not originate. The persistence law is a
property of integrated ordered context, not a sum of separable span contributions: distant
material exerts its influence as part of the ordered whole rather than through its internal
arrangement alone.

\subsection{Persistence is broadly distributed across prior sentences}

We next asked whether persistence is concentrated in a small number of privileged prior
sentences or broadly distributed across the prior context. We addressed this with a
sentence-level ablation: for each target, we removed one prior sentence at a time and
measured the resulting change in prediction.

\begin{figure}[t]
\centering
\includegraphics[width=0.85\textwidth]{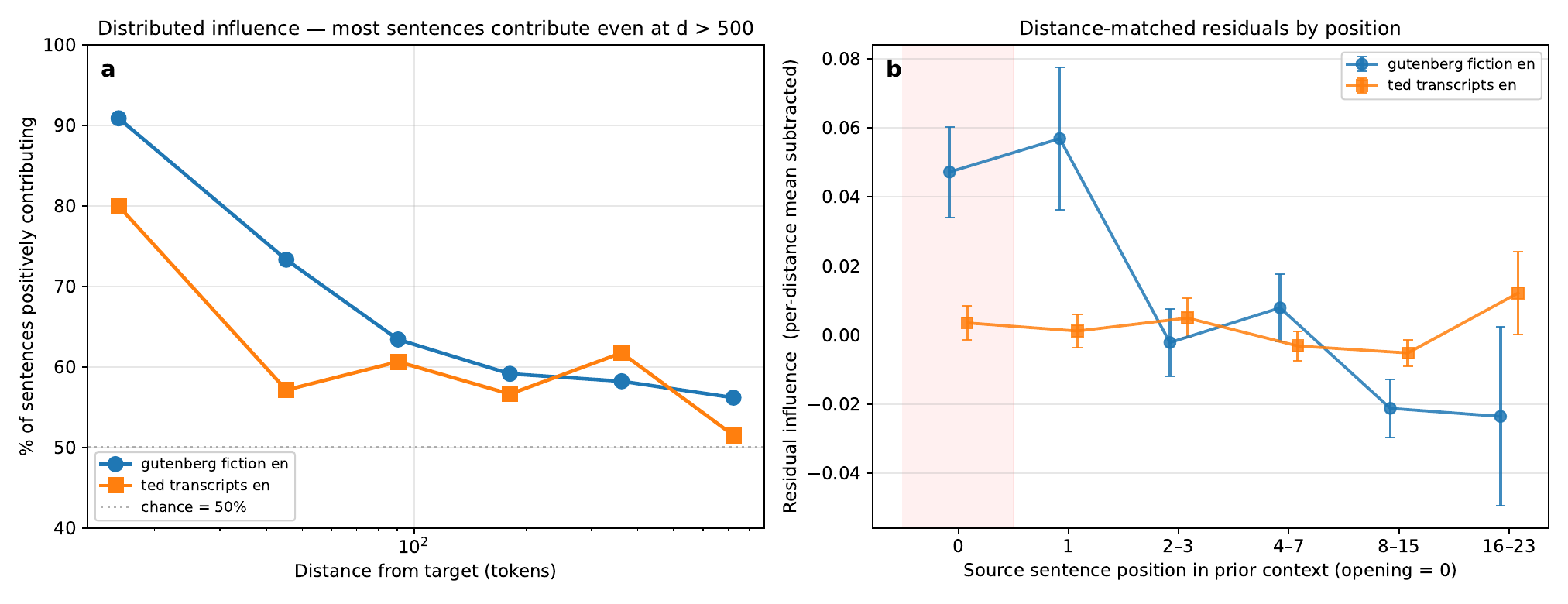}
\caption{\textbf{Persistence is broadly distributed across prior sentences.} \textbf{a,} Proportion of prior
sentences whose removal degrades target prediction, as a function of distance to the target. The
proportion declines smoothly from ${\sim}90\%$ within the first 32 tokens to ${\sim}55$--$60\%$ beyond
500 tokens, with no sharp transition. The top-5 sentences account for 30--51\% of total positive
influence (Gini 0.78--0.90), but this concentration reflects recency, not a privileged structural role
for any specific position. \textbf{b,} Mean per-sentence influence by absolute position in the prior context, after
subtracting the mean influence at each exact sentence distance (nonparametric distance control;
leave-one-sentence-out influence matrices, 15 documents per corpus). Positions 0 and 1 are shown
individually; later positions are aggregated in dyadic bins (2--3, 4--7, 8--15, 16--23). Shaded
band marks the opening position; error bars, s.e.m. TED residuals are flat across positions. Fiction
shows a small elevation at the earliest two positions (${\sim}0.05$, an order of magnitude below
its mean per-sentence influence), consistent with the added cost of removing sentences that
introduce recurring referents. No position carries influence beyond distance on a scale comparable
to the distance effect itself: influence is a distributed field graded by recency, not anchored at
privileged positions.}
\label{fig:distributed}
\end{figure}

The majority of prior sentences contribute positively to prediction at all distances
examined. The proportion of contributing sentences declines gradually with distance,
from approximately 90\% at short range to 55--60\% beyond 500 tokens, with no sharp
transition or threshold (Figure~\ref{fig:distributed}). The magnitude of contribution is concentrated at
recent positions (top-5 sentences account for 30--51\% of total positive influence;
Gini coefficient 0.78--0.90), but this concentration is attributable to recency.
After matching for distance exactly, influence by position is flat in TED transcripts; fiction
shows one small, interpretable exception --- a ${\sim}0.05$ elevation at the earliest one to two
positions, an order of magnitude below mean per-sentence influence, consistent with the cost of
removing sentences that introduce recurring referents (Figure~\ref{fig:distributed}b). Persistence is therefore a distributed field of influence across the
prior sequence, with magnitude graded smoothly by recency rather than localized to a
small number of anchoring positions.

\section{Discussion}

Our observations establish a scaling law, which we term the
contextual persistence law: contextual influence in human language decays as an approximately
scale-free power law, with a near-constant exponent across ten corpora, six language families,
and both written and spoken modalities. Because that exponent is close to $1$, contextual
influence is distributed approximately uniformly across logarithmic timescales rather than
concentrated at any one. Unlike generic heavy-tailed behaviour, an exponent near unity occupies
the unique boundary at which aggregate influence per logarithmic decade is approximately constant
--- for $\alpha>1$ influence concentrates at short range, for $\alpha<1$ at long range --- so the
law specifies not merely the absence of a characteristic scale but a particular organization of
influence across scales. The
consistency of the exponent across such diverse samples makes this a candidate
cross-linguistic regularity of human language, in the same empirical sense that Zipf's and Heaps'
laws are cross-linguistic regularities. In the hierarchy of statistical descriptions considered here ---
unit frequency (Zipf's law), pairwise dependence across distance (long-range mutual information),
and order-specific dependence --- the contextual persistence law is the third, order-specific rung.
Where previous scaling laws describe distributions of units or pairwise relations among them, this
law characterizes the scaling of order-specific contextual influence: the sequential organization
of predictive dependence across distance, the component that remains once unit identity
and co-occurrence are removed. It is a law to be
explained rather than an adjudication between mechanisms: any account of how
human language is generated must explain why its statistical product lacks a detectable
characteristic scale over this range. An architecture with a fixed maintenance
horizon would generally be expected to produce an inflection in $P(d)$ near that horizon, of
whatever size; we observe none across nearly two orders of magnitude. An account in which long-range coherence rests on a small
number of privileged anchor positions would generally predict a concentrated rather than a
distributed influence field; we observe broad, recency-graded distribution. And an account in
which distant material contributes through the internal structure of isolated episodes would
predict that the law could be reconstructed from span-internal order; the increment-shuffle
experiment shows it cannot: band-local order decays too steeply, and the scale-free law emerges
only at the level of integrated ordered context. We do not take
these observations to settle how memory is organized during production; they characterize
the product, not the process. But the law they describe is what every proposed process
must reproduce.

Why should such a law exist? The mean exponent we report ($\alpha=1.04$, SD $=0.15$) sits in a
neighborhood independently identified in memory research by methods that involve no language
model: the odds that a memory is needed in natural environments \citep{anderson1991reflections},
associative decay in free recall \citep{kahana2002age}, and the form of forgetting across human
and animal tasks \citep{wixted1991form} all follow power functions with order-unity exponents. We
do not claim a single mechanism generates them all, but the recurrence raises the possibility (a
hypothesis, not a finding) that the persistence law reflects the same constraints operating in
human language generation: the accessibility of prior content during production falling off in
the same graded, scale-free way that memory itself decays. If this account is accurate, it
dissolves a familiar discontinuity. Architectures that divide memory into a capacity-limited
store for the present and a separate long-term store for the past should leave some boundary
signature in the product, and none appears: the law is continuous across the range where any such
boundary would fall, consistent with graded rather than architecturally bounded characterizations
of memory \citep{cowan2001magical, ericsson1995longterm, ma2014changing}. The observation doing the work
remains the same: a near-$1/d$ persistence law whose near-unity exponent places language at the
boundary where each logarithmic timescale contributes approximately equally to prediction.

The CPF characterizes the statistical structure of language, not directly the memory process
that produced it: it is computed over text, and predictive structure in text need not imply that
speakers and writers actively used long-range memory to generate it. The property is nonetheless
far more constrained than a generic statistical regularity: because $P(d)$ subtracts a
shuffled-token baseline, vanishes on token-scrambled and frequency-matched synthetic sequences,
and replicates across independently trained probes, the long-range dependence reflects genuine
order-dependent structure in what humans produce, not lexical co-occurrence or measurement bias. A related objection, that written text is a static
visual trace an author can edit and cross-reference offline unlike online cognition, is
tempered by the modalities in which the law holds: the same signature appears in spontaneous
conversational speech, where offline revision is impossible, and in the spoken delivery of
prepared TED discourse. The observation across both spontaneous and prepared speech indicates
that the law is not contingent on written revision.

A parsimonious reading of these results is that the human generative process is itself
autoregressive: each word is produced against the accumulated ordered history of the sequence,
and the persistence law is the trace that this mechanism leaves in its product. On this reading
the probe is not an arbitrary instrument but one matched in form to the process it measures. We
advance this as an interpretation, not a demonstrated mechanism: the measurements characterize
the product, and other generative architectures --- hierarchical planning, retrieval over
structured discourse representations --- could conceivably yield the same scale-free profile.
Nor can modern language models decide the question: they are explicitly autoregressive, but they
are trained on human text and inherit its statistics, so their reproduction of the law shows only
that they have learned it. What would adjudicate is evidence from the human process itself.
Converting the present result from a property of the product into a claim about the producer,
or the comprehender, will require online behavioural and neural evidence, and the law makes
this directly testable because it predicts a specific quantity. The central prediction is that the
cost of disrupting prior context should itself scale as $P(d)$: shuffling or replacing context in
log-spaced windows (for example, tokens 10--20 back versus 100--110 back) should slow human
reading and lengthen gaze durations on a downstream target \citep{hale2001probabilistic, levy2008expectation, smith2013effect, futrell2020lossy} by amounts that fall off with distance
at the near-unity rate, rather than dropping sharply at a working-memory horizon. Parallel
predictions follow for proactive interference during comprehension and for the context-driven
reduction of the N400 \citep{kutas2011thirty}, each expected to follow the same
$\alpha \approx 1$ decay; we regard adjudicating these as the key open hypothesis the law
motivates.

Human language exhibits a simple quantitative law, the contextual
persistence law, in which the influence of prior context is distributed approximately
uniformly across logarithmic distances in human-produced language. Whether this regularity reflects
properties of memory, discourse organization, planning, prediction, or communication is
precisely the mechanistic question the law leaves open; whether it generalizes
to non-linguistic sequential behaviours with extended temporal coherence (music improvisation,
motor sequences, drawing) remains to be tested. But like other scaling laws in language, its
value lies less in any mechanism it immediately reveals than in the constraint it imposes on
every mechanism proposed thereafter: its existence is now an empirical target that any theory of
language must account for.

\section{Methods}

\subsection{Corpora}

The long-range CPF was computed on ten corpora: English fiction (public-domain texts
from Project Gutenberg, \url{https://www.gutenberg.org}), English news prose (the English/BBC
subset of XL-Sum \citep{hasan2021xlsum}), English spontaneous speech (Buckeye Corpus of Conversational
Speech \citep{pitt2007buckeye}), TED transcripts in English, German, French, Turkish, and Russian
(OPUS TED2020 \citep{reimers2020making, tiedemann2012parallel}), and literary prose in Japanese
(Aozora Bunko) and Finnish (Project Gutenberg). Together these cover six language families (Germanic, Romance, Turkic,
Slavic, Japonic, Uralic) and four production modes (written fiction, news prose, prepared
speech, spontaneous speech). Per-corpus document counts (documents contributing at least one
full-length target window) were: English fiction 60, English news 60, Buckeye 26, TED English 60,
German 60, French 31, Turkish 12, Russian 60, Japanese literary 42, Finnish literary 38 --- 449
documents in total. The complementary cross-language qualitative replication of the sign-flip
pattern (Afro-Asiatic, Koreanic, Sino-Tibetan) was performed on Wikipedia and TED-transcript cells
from a broader corpus inventory at shorter context lengths. Full per-corpus provenance --- sources,
identifiers, licenses, snapshot dates, and selection criteria --- is given in Supplementary Note~5 (Table~S3), and preprocessing steps are described there.

\subsection{Probe models}

Two autoregressive language models served as probes: Llama-3.1-8B (base;
\texttt{Meta-Llama-3.1-8B}) \citep{llama3herd2024} and Mistral-7B-v0.1
(\texttt{mistralai/Mistral-7B-v0.1}) \citep{jiang2023mistral}. Neither was fine-tuned for this study. The
probes were used only to compute conditional probability estimates over fixed token sequences;
the language models themselves were not the object of study. Headline results were obtained on
Llama-3.1-8B; the qualitative pattern was replicated on Mistral-7B for shared corpora as a
probe-independence check.

\subsection{Contextual persistence function}

For each document, a single target region of 30 tokens was selected at the midpoint (50\%) of
document length. Context preceding the target was revealed in incrementally larger spans at
log-spaced context lengths $c \in \{0, 1, 2, 4, 8, 16, 32, 64, 128, 256, 512, 1024\}$. At each
context length, target perplexity was computed under intact context (the actual preceding $c$
tokens, original order) and shuffled context (the same $c$ tokens, random order). The shuffled
baseline used a single random permutation of the prior-context tokens at each target and context
length, drawn with a fixed seed so that the baseline is deterministic and reproducible and the same
permutation is applied across probe models; the 30-token target itself was never permuted, and no
sentence-boundary or special tokens were introduced into the shuffled span. Sensitivity of the
fitted exponent to this single-permutation choice is quantified in Supplementary Note~3, where the analysis is
repeated with $K=20$ independent permutations per target and context length. Per-distance
marginals were obtained as the per-token change in perplexity across successive context lengths,
$m_d = (\mathrm{ppl}_{c_{i-1}} - \mathrm{ppl}_{c_i}) / (c_i - c_{i-1})$, with distance indexed by
the geometric mean $d = \sqrt{c_{i-1} c_i}$ of the interval. The CPF was defined
as $P(d) = m_d^{\text{intact}} - m_d^{\text{shuffled}}$. The shuffle preserves the identity and
composition of the prior context while destroying its order, so this subtraction separates the
predictive benefit of intact order from the benefit available from the same token composition in
scrambled order, and from probe-internal token-frequency calibration. Marginals were aggregated across
documents within each corpus and binned by distance. Only documents with at least
$1024 + 30 + 50$ tokens were included, ensuring a full-length context window and target for every
measurement.

\subsection{Functional-form fitting}

The per-token order-specific gap was fit to five candidate functional forms (power law,
exponential, stretched exponential, truncated power law, and quadratic-in-log,
i.e.\ log-normal-in-distance) and ranked per corpus by the small-sample-corrected Akaike
criterion (AICc); the full comparison is reported in Supplementary Note~1. Fits were performed
in log space on the fit range $d \geq 10$ tokens, the first interval at which both ordered and
shuffled bases are at least eight tokens; shorter intervals were excluded a priori because with
only one or two context tokens there is no meaningful permutation, leaving the shuffled baseline,
and hence $P(d)$, unstable at $d < 10$. Within this fixed fit range the aggregated (corpus-mean)
$P(d)$ is positive at every distance for all ten language corpora, so no bins were dropped from any
language fit; the positive-value requirement is therefore not a source of tail-selection bias for
the headline result and becomes operative only for the non-language controls, where sign changes
occur and we report the number of positive bins used (Figure~\ref{fig:langbio}).

\subsection{Synthetic-sequence controls}

To distinguish probe-internal calibration dynamics from genuine sequential dependence, two
synthetic-sequence controls were constructed: uniform-vocabulary sequences (tokens drawn uniformly
at random from the model vocabulary) and frequency-matched sequences (tokens drawn with empirical
token-frequency probabilities matched to the natural corpora). The same CPF
protocol was applied to both.

\subsection{Non-language domain controls}

To test whether the persistence law is specific to human language or a generic property of
autoregressive probing over any structured sequence, the identical CPF protocol was applied
to two non-language biological-sequence domains, each with a domain-native autoregressive
probe. For DNA, the probe was HyenaDNA \citep{nguyen2023hyenadna}, a single-nucleotide
autoregressive genomic language model, run at two scales
(\texttt{hyenadna-medium-160k-seqlen} and \texttt{hyenadna-large-1m-seqlen}); sequences were
non-overlapping ACGT windows from human reference chromosome~21 (GRCh38, Ensembl
\citep{cunningham2022ensembl}). For protein, the
probe was ProGen2 \citep{nijkamp2023progen2}, an autoregressive decoder protein language
model, run at two scales (\texttt{progen2-small}, 151M; \texttt{progen2-large}, 2.7B);
sequences were UniProtKB/Swiss-Prot proteins \citep{uniprot2023} restricted to the twenty standard amino acids. In both
domains the protocol matched the language runs: a 30-residue target at the sequence midpoint,
log-spaced context lengths capped at the model's context window, an intact-versus-shuffled
contrast, and power-law fitting on positive bins at $d \geq 10$. Because these ports are
custom, probe faithfulness was verified before analysis: held-out per-residue perplexity was
reproduced (ProGen2: 15.1 and 15.7 for the small and large models respectively, with per-protein
spread and well below the $\approx 20$ uniform-amino-acid baseline), and a short-range
order-specific floor was confirmed. For protein, the exponent was estimated on a dense
log-spaced grid (approximately 28 bins between $d = 10$ and $993$) over 300 documents at each
scale, with 95\% confidence intervals from a 1{,}000-fold document bootstrap using the same
estimator applied to the language corpora. Full domain-control results and diagnostics are
reported in Supplementary Information.

\subsection{Sentence-level decomposition}

The effects of sentence-order reversal, discourse-sequence permutation, and within-sentence content were
assessed by computing $P(d)$ under four context conditions on the same target windows: ordered
(intact); sentence-shuffled (the prior context segmented at sentence boundaries and the sentences
randomly permuted, with within-sentence token order preserved); sentence-reversed (the order of the
sentences reversed, so that the sentence nearest the target is moved farthest from it, with
within-sentence token order preserved), which preserves the set of sentence adjacencies while
reversing their order and target-relative positions; and token-shuffled (all context tokens randomly
permuted). From these we formed three contrasts, each fit by least squares on log--log axes over the
positive bins at $d \geq 10$: the sentence-reversal contrast (ordered minus sentence-reversed), which
measures the loss from reversing the temporal organization of the prior discourse; the
sequence-permutation contrast (ordered minus sentence-shuffled), which isolates sentence-sequence
organization; and the sentence-content contrast (sentence-shuffled minus token-shuffled), which
estimates the contribution retained by sentence content. Slopes are reported
as per-cell means $\pm$ SD across cells. The decomposition was computed across eight cells: the
long-range corpora excluding Buckeye (spontaneous speech lacks unambiguous sentence boundaries) and
Russian (added to the headline analysis after the decomposition run). Figure~\ref{fig:sentshuf}
aggregates these eight cells.

\subsection{Sentence-level ablation}

For documents in two corpora (English fiction and English TED transcripts, 20 documents per corpus,
1024-token prior context), each prior sentence was individually removed and the resulting change in
target perplexity was computed (leave-one-sentence-out). Per-sentence influence was characterized by
sign (positive $=$ removing the sentence increased target perplexity) and magnitude. The proportion
of contributing sentences was computed in distance bins, and concentration metrics (Gini coefficient
over per-sentence magnitudes; share of total positive influence in the top-5 sentences) were computed
per document and aggregated. To test whether magnitude variation across prior-sentence positions
reflects a privileged structural role over and above distance from the target, the full
leave-one-sentence-out influence matrix was computed on a 15-document subset of each ablation corpus
(standardized 25-sentence prior-context windows). For the positional analysis, per-sentence
influence values were demeaned within each exact sentence distance --- a nonparametric distance
control that imposes no functional form and requires no exclusions --- and mean residuals were
examined by absolute position, with positions 0 and 1 reported individually and later positions
aggregated in dyadic bins (2--3, 4--7, 8--15, 16--23) for stable estimates.
Distance-matched residuals are flat across positions in TED transcripts; in fiction the earliest
one to two positions show a small positive residual (${\sim}0.05$ per sentence, an order of
magnitude below mean per-sentence influence), consistent with the reintroduction cost of referents
established at the opening. No structural position carries influence beyond distance on the scale
of the distance effect itself.

\section*{Supplementary Information}

\subsection*{Supplementary Note 1: functional-form robustness}

Because scale-free claims are sensitive to the choice of fitting procedure and to
competing functional forms, we treated functional-form selection as a robustness analysis
rather than assuming a power law. For each corpus the per-token order-specific gap was fit,
in log space, to five candidate forms: a power law ($\log y = a + b\log d$), an exponential
($\log y = a + bd$), a stretched exponential ($\log y = a - \beta d^{\gamma}$), a truncated
power law ($\log y = a + b\log d - d/\lambda$), and a quadratic-in-log (log-normal-in-distance)
form ($\log y = a + b\log d + c(\log d)^2$). Models were ranked per corpus by the small-sample
corrected Akaike criterion (AICc), which penalizes the additional free parameters of the more
flexible forms; each fit used the seven binned points with $d \geq 10$ and positive influence.

The power law is preferred in eight of the ten corpora (Table~S1) and outperforms the pure
exponential (the canonical fixed-scale alternative) in all ten, by $\Delta\text{AICc}$
ranging from $+0.3$ to $+16.8$ (median $\approx +14$); the exponential is never preferred. The
truncated power law never improves on the plain power law: in six corpora its fitted decay term
is non-negative (no cutoff at all), and in three of the remaining four the implied cutoff
$\lambda$ lies beyond the measured range. Only Buckeye spontaneous speech shows a truncated-power-law
cutoff within range ($\lambda \approx 276$ tokens). Separately, and as a distinct model comparison,
Buckeye is also the one corpus for which the stretched exponential is marginally preferred over the
power law by AICc, while Japanese literary prose marginally prefers the quadratic-in-log form. Both the
cutoff and the stretched-exponential preference point to mild downward curvature at the longest distances
in this cell, but they are separate tests and should not be read as one explaining the other. In both non-power-law cases the power law remains a
close second (margin $<2$ AICc), the fitted stretched exponent stays well below one, and the decay
remains heavy-tailed. The fitted power-law exponents
are tightly clustered near $1$ (range $\alpha=0.75$ to $1.31$; eight of ten within $0.93$ to $1.20$).
The maximum-likelihood, Kolmogorov--Smirnov procedure of \citet{clauset2009power} addresses
whether a \emph{sample} is drawn from a heavy-tailed distribution rather than the present
setting of regressing a response, $P(d)$, on distance; we therefore compared candidate decay
functions using AICc. No alternative form removes the central observation: across the measured
range the decay carries no characteristic scale.

\renewcommand{\thetable}{S1}
\begin{table}[h]
\centering
\small
\begin{tabular}{lcccc}
\toprule
Corpus & $N$ & Preferred form & $\Delta$AICc (exp $-$ power) & $\alpha$ [95\% CI] \\
\midrule
Gutenberg (fiction, en) & 60 & power law & $+12.9$ & $1.05$ [$0.90$, $1.33$] \\
TED (en)                & 60 & power law & $+15.5$ & $0.97$ [$0.87$, $1.11$] \\
TED (de)                & 60 & power law & $+15.5$ & $1.03$ [$0.93$, $1.16$] \\
TED (fr)                & 31 & power law & $+9.4$  & $1.31$ [$1.02$, $1.57$] \\
TED (tr)                & 12 & power law & $+5.0$  & $0.75$ [$0.37$, $1.02$] \\
TED (ru)                & 60 & power law & $+16.8$ & $1.02$ [$0.94$, $1.13$] \\
Literary prose (ja)     & 42 & quadratic-in-log & $+13.2$ & $0.98$ [$0.84$, $1.15$] \\
Literary prose (fi)     & 38 & power law & $+14.9$ & $0.93$ [$0.77$, $1.07$] \\
News (en)               & 60 & power law & $+13.6$ & $1.16$ [$0.98$, $1.37$] \\
Buckeye (speech, en)    & 26 & stretched exp. & $+0.3$ & $1.20$ [$0.82$, $1.53$] \\
\bottomrule
\end{tabular}
\caption{Functional-form comparison and per-corpus exponents across the ten long-range corpora
(Table~S1). $N$ is the number of contributing documents. $\Delta$AICc (exp $-$ power) is positive
whenever the power law is preferred over a pure exponential; it is positive in every corpus, and the
power law is the AICc-preferred form in eight of ten (the two exceptions show mild downward curvature,
with the power law a close second). Exponent 95\% confidence intervals are from a document-level
bootstrap ($B=2000$); the wider intervals correspond to the smaller corpora (Turkish, Buckeye, French).}
\end{table}

\subsection*{Supplementary Note 2: robustness of the exponent --- units and heterogeneity}

Two further checks bound the interpretation of the exponent. First, because $P(d)$ is defined as a
difference of perplexities, and perplexity is exponential in the underlying log-loss, we recomputed
the entire analysis with $P(d)$ defined in log-probability (nats) units (i.e., replacing each
perplexity by its logarithm, an additive information quantity) and refit the power law. The
scale-free \emph{form} is robust to this choice: every corpus remains a near-$1/d$ power law and the
ordering of corpora is preserved. The precise exponent, however, is not unit-invariant: the mean
shifts from $\alpha=1.04$ (perplexity) to $\alpha=1.18$ (log-probability), a consistent offset of ${\approx}\,0.14$
present in every cell. We report perplexity-based values in the main text for continuity with the raw
measurement and note that the order-unity exponent, and the associated approximate uniformity of
influence across logarithmic timescales, hold in both unit conventions, with the exact value
dependent on the choice.

Second, a document-level bootstrap ($B=2000$ resamples of documents within each corpus) yields the
per-corpus confidence intervals in Table~S1. The between-corpus standard deviation of the ten point
estimates ($0.156$) modestly exceeds the typical within-corpus bootstrap standard error
(${\approx}\,0.10$). The permutation-sensitivity analysis below shows that part of this residual
spread reflects noise in the single-permutation shuffled baseline rather than genuine linguistic
variation: averaging over permutations narrows the apparent cross-corpus range. The exponent is
therefore best described as clustered near $1$, with limited evidence for residual corpus-level
heterogeneity and the low-$N$ corpora (Turkish, Buckeye, French) contributing most of both the
apparent spread and the widest intervals.

\subsection*{Supplementary Note 3: robustness of the exponent --- sensitivity to the shuffled permutation}

Because $P(d)$ subtracts a shuffled-context baseline, and the main analysis used a single random
permutation of the prior context at each target and context length, we tested whether the fitted
exponent depends on that choice with a permutation-seed sensitivity analysis. On a subset of up to
$25$ documents per corpus (all $26$ for Buckeye), we redrew $K=20$ independent permutations at each
target and context length and recorded the target perplexity under every one, so that both the
permutation-averaged baseline and the full across-permutation distribution of the exponent are
available; permutation index $0$ reused the original seed. Three results follow. First, the
cross-corpus mean exponent is unchanged: averaging the shuffled baseline over the $20$ permutations
gives a mean $\alpha=1.04$, and the mean of the $20$ individual single-permutation estimates is
$\alpha=1.05$, both matching the single-permutation value of $1.04$ in the main text. Second,
averaging the baseline improves the power-law fits: corpora whose single-permutation fits were
degraded by shuffle noise rise to $r^2\approx0.97$ once the averaged baseline removes per-permutation
fluctuation in the shuffled perplexity (TED German $r^2\,0.66\!\to\!0.97$; Finnish literary
$0.55\!\to\!0.97$; TED French $0.82\!\to\!0.97$; TED Turkish $0.86\!\to\!0.98$). Third, the per-corpus
exponent carries non-negligible permutation noise, concentrated in the smaller corpora: the
across-permutation standard deviation of $\alpha$ reaches ${\approx}\,0.20$ (Finnish literary prose,
English news) while the larger corpora vary by ${<}\,0.1$, and a single permutation can differ from
the $20$-permutation average by up to ${\approx}\,0.28$ (TED French). Averaging accordingly narrows
the apparent cross-corpus spread, indicating that part of the between-corpus heterogeneity in
Table~S1 reflects permutation noise rather than genuine linguistic variation. No corpus-level
conclusion depends on the chosen permutation, and the near-unity, scale-free signature is unchanged.
Per-corpus exponents under the single-permutation and $K=20$ averaged baselines are given in
Table~S2.

\renewcommand{\thetable}{S2}
\begin{table}[h]
\centering
\small
\begin{tabular}{lcccc}
\toprule
Corpus & $\alpha$ (single perm.) & $\alpha$ (avg., $K=20$) & $r^2$ (avg.) & SD$_\alpha$ (perms) \\
\midrule
Gutenberg (fiction, en) & $1.15$ & $1.10$ & $0.93$ & $0.09$ \\
TED (en)                & $1.04$ & $0.97$ & $0.96$ & $0.09$ \\
TED (de)                & $1.09$ & $1.07$ & $0.97$ & $0.11$ \\
TED (fr)                & $1.43$ & $1.16$ & $0.97$ & $0.12$ \\
TED (tr)                & $0.75$ & $0.90$ & $0.98$ & $0.07$ \\
Literary prose (ja)     & $0.92$ & $0.99$ & $0.98$ & $0.09$ \\
Literary prose (fi)     & $0.71$ & $0.87$ & $0.97$ & $0.20$ \\
News (en)               & $1.22$ & $1.19$ & $0.95$ & $0.18$ \\
Buckeye (speech, en)    & $1.20$ & $1.11$ & $0.95$ & $0.16$ \\
\bottomrule
\end{tabular}
\caption{Permutation-seed sensitivity of the per-corpus decay exponent (Table~S2), Llama-3.1-8B
probe, on a subset of up to $25$ documents per corpus (all $26$ for Buckeye). ``Single perm.'' reuses
the original seed; ``avg., $K=20$'' averages the shuffled baseline over twenty independent
permutations per target and context length; SD$_\alpha$ is the standard deviation of the exponent
across the twenty permutations. Averaging leaves the cross-corpus mean at $\alpha=1.04$ while raising
the poorer single-permutation fits to $r^2\approx0.97$. The Russian TED cell, computed with a
separate pipeline in the main analysis, was not included in this check.}
\end{table}

\subsection*{Supplementary Note 4: robustness of the exponent --- sensitivity to the fit range}

Headline fits use distance bins with $d \geq 10$ tokens, excluding the shortest intervals, where
the shuffled-token baseline is unstable (Methods). Because context lengths are log-spaced,
candidate cutoffs quantize onto the interval midpoints ($d \approx 5.7, 11.3, 22.6, \ldots$):
cutoffs of 8 and 10 tokens select identical bin sets, as do 12 and 16. Refitting every corpus at
each cutoff, the headline range ($d \geq 8$ or $10$) yields mean $\alpha = 1.04$ (SD $= 0.15$);
additionally excluding the $d \approx 11.3$ interval ($d \geq 12$ or $16$) leaves the mean
unchanged ($\alpha = 1.04$, SD $= 0.21$); and including the $d \approx 5.7$ interval
($d \geq 5$) shifts it to $\alpha = 0.89$ (SD $= 0.10$), reflecting the documented short-range
instability of the baseline rather than a property of the decay. The exponent estimate is
therefore insensitive to the cutoff within the stable range, and its only sensitivity is to
intervals excluded on independent grounds.

\subsection*{Supplementary Note 5: corpus sources and provenance}

Table~S3 lists the source, license, and snapshot or version for each of the ten long-range corpora;
preprocessing and selection are summarized here. English fiction comprises public-domain novels from
Project Gutenberg (catalogue \texttt{pg\_catalog.csv}; plain-text ebooks), selected as English fiction
by Library of Congress class (PR, PS, PZ) or fiction subject headings, ordered by ascending Gutenberg
ID to favour canonical works, with a per-author cap and rejection of verse, drama, and OCR artefacts,
then segmented into ${\approx}3{,}000$-word chunks. English news is the English (BBC) portion of XL-Sum
\citep{hasan2021xlsum}. English spontaneous speech is the Buckeye Corpus \citep{pitt2007buckeye,
pitt2005buckeye}, retaining interviewee turns. TED transcripts (English, German, French, Turkish,
Russian) are from the OPUS TED2020 monolingual release \citep{reimers2020making, tiedemann2012parallel}
(version v1), retaining talks of at least $900$ words after a per-language junk-ratio filter, one
segment per talk. Japanese literary prose comprises public-domain texts from Aozora Bunko (Nippon
Decimal Classification 913/914/915; 21 authors, 42 texts) and Finnish literary prose public-domain
texts from Project Gutenberg (19 authors, 38 texts); both were captured in a snapshot dated
2026-05-01. The cross-language qualitative replication (Afro-Asiatic, Koreanic, Sino-Tibetan) drew
additionally on Wikipedia (dumps current at the time of analysis; CC BY-SA 3.0), sampled at a
500-token minimum after removing markup, references, and infoboxes. Tokenization used each probe's
native tokenizer throughout.

\renewcommand{\thetable}{S3}
\begin{table}[h]
\centering
\footnotesize
\begin{tabular}{p{3.0cm}p{4.6cm}p{2.6cm}p{2.6cm}}
\toprule
Corpus & Source (reference) & License & Snapshot / version \\
\midrule
English fiction        & Project Gutenberg                                            & Public domain (US) & 2026 snapshot \\
English news           & XL-Sum, BBC subset \citep{hasan2021xlsum}                    & See source         & \texttt{csebuetnlp/xlsum} \\
English speech         & Buckeye \citep{pitt2007buckeye, pitt2005buckeye}             & Free for research  & 2007 release \\
TED (en, de, fr, tr, ru) & OPUS TED2020 \citep{reimers2020making, tiedemann2012parallel} & See source         & v1 \\
Literary prose (ja)    & Aozora Bunko                                                 & Public domain      & 2026-05-01 \\
Literary prose (fi)    & Project Gutenberg                                            & Public domain (US) & 2026-05-01 \\
Wikipedia (replication) & Wikimedia dumps                                             & CC BY-SA 3.0       & Dump at analysis time \\
\bottomrule
\end{tabular}
\caption{Sources and provenance for the language corpora (Table~S3). ``See source'' indicates the
distributing repository states the terms of use; no separate license file was recorded at ingestion.
Snapshot dates are given where recorded; the Project Gutenberg fiction set was captured in the same
2026 collection window but without a per-file date stamp.}
\end{table}

\renewcommand{\thefigure}{S1}
\begin{figure}[h]
\centering
\IfFileExists{figures/exp2_qd_loglog.pdf}{%
  \includegraphics[width=\textwidth]{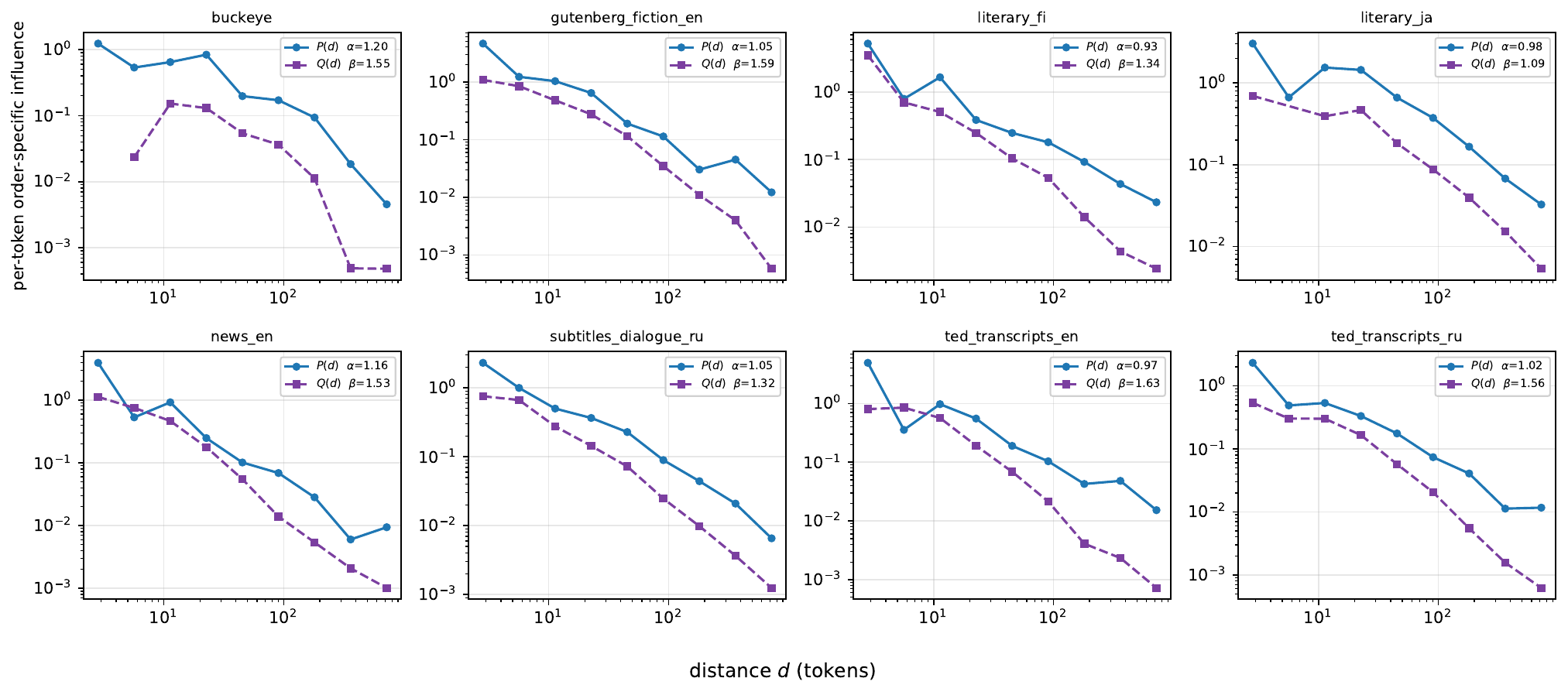}}{%
  \fbox{\parbox[c][6cm][c]{0.9\textwidth}{\centering \texttt{figures/exp2\_qd\_loglog.pdf} --- generate
  with \texttt{analysis/exp2\_qd\_figure.py} after placing the Exp 2 caches in
  \texttt{results/exp2\_increment\_shuffle/llama/}.}}}
\caption{\textbf{Span-internal order contribution $Q(d)$ and range-matched marginal persistence
$P(d)$, per corpus.} For each of the eight corpora with both estimates (Llama-3.1-8B), the
increment-shuffle band-local order function $Q(d)$ and the marginal persistence function $P(d)$ on
log--log axes. $Q(d)$ is positive throughout but steeper: the band-local exponent $\beta$ exceeds the
range-matched integrated exponent $\alpha$ in every corpus. Per-corpus fits and values are in
\texttt{results/exp2\_increment\_shuffle/qd\_summary.csv}.}
\label{fig:qd}
\end{figure}

\section*{Data and code availability}

Analysis code, corpus manifests, per-corpus persistence curves, and the figure-generation pipeline
are available at \url{https://github.com/elanbarenholtz/contextual-persistence}. An archived version with an
assigned DOI will accompany journal publication. Corpora that cannot be redistributed under their
source licenses (notably the Buckeye Corpus) must be obtained from the original providers.

\section*{Author contributions}

E.B. conceived and designed the study, performed the analyses, and wrote the manuscript.

\section*{Funding}

The author received no specific funding for this work.

\section*{Competing interests}

The author declares no competing interests.

\bibliographystyle{plainnat}
\bibliography{references}

\end{document}